\documentclass[11pt,a4paper]{article}
\usepackage[hyperref]{structured-prediction}
\usepackage{url}
\usepackage{latexsym}
\usepackage{graphicx}
\usepackage{hyperref}
\usepackage{times}
\usepackage{xcolor}
\usepackage{latexsym}
\usepackage{csquotes}
\usepackage{multicol,array,booktabs,siunitx,multirow,dcolumn}
\usepackage{tikz-dependency,subcaption,float}

\renewrobustcmd{\bfseries}{\fontseries{b}\selectfont}
\usepackage{microtype}
\aclfinalcopy 

\definecolor{deppink}{HTML}{a92b7a}
\definecolor{depgreen}{HTML}{679033}
\definecolor{deporange}{HTML}{be6320}
\definecolor{depblue}{HTML}{165f77}
\definecolor{depred}{HTML}{b12019}

\newcommand{\invzero}{\textcolor{white}{0}}

\title{\textit{A Falta de Pan, Buenas Son Tortas:}\thanks{\hspace{0.5em}\textit{Lacking yeast-proven bread, a flatbread alternative will suffice}, i.e. if you can't get more fully-annotated dependency trees, annotating UPOS tags can still be helpful.}\\The Efficacy of Predicted UPOS Tags for Low Resource UD Parsing}
\author{Mark Anderson \\
 Universidade da Coru\~na, CITIC\\
 Department of CS \& IT \\
  {\tt m.anderson@udc} \\\And
  Mathieu Dehouck\\
  LaTTiCe, CNRS, ENS\\
  Université Sorbonne Nouvelle\\
  {\tt mathieu.dehouck@udc.es} \\ \And
    Carlos G{\'o}mez-Rodr{\'\i}guez \\
 Universidade da Coru\~na, CITIC\\
 Department of CS \& IT \\
  {\tt carlos.gomez@udc.es}}

\date{}
\definecolor{deppink}{HTML}{a92b7a}
\definecolor{depgreen}{HTML}{679033}
\definecolor{deporange}{HTML}{be6320}
\definecolor{depblue}{HTML}{165f77}
\definecolor{depred}{HTML}{b12019}
\definecolor{depgrey}{HTML}{6c8093}
\definecolor{white}{HTML}{ffffff}

\makeatletter

\newcommand*{\@rowstyle}{}

\newcommand*{\rowstyle}[1]{
  \gdef\@rowstyle{#1}%
  \@rowstyle\ignorespaces%
}

\newcolumntype{=}{
  >{\gdef\@rowstyle{}}%
}

\newcolumntype{+}{
  >{\@rowstyle}%
}

\makeatother

\begin{document}
\maketitle
\begin{abstract} 
 We evaluate the efficacy of predicted UPOS tags as input features for dependency parsers in lower resource settings to evaluate how treebank size affects the impact tagging accuracy has on parsing performance. We do this for real low resource universal dependency treebanks, artificially low resource data with varying treebank sizes, and for very small treebanks with varying amounts of augmented data. We find that predicted UPOS tags are somewhat helpful for low resource treebanks, especially when fewer fully-annotated trees are available. We also find that this positive impact diminishes as the amount of data increases.
\end{abstract}
\section{Introduction}
Low resource parsing is a long-standing problem in NLP and many techniques have been introduced to tackle it \cite{hwa2005bootstrapping,zeman2008cross,ganchev2009dependency,mcdonald2011multi,agic2016multilingual}. For an extensive review and comparison of techniques see \citet{vania2019a}. Here we focus on the utility of part-of-speech (POS) tags as features for low resource dependency parsers.

POS tags are a common feature for dependency parsers. \citet{tiedemann2015cross} highlighted the unrealistic performance of low resource parsers when using gold POS tags in a simulated low resource setting. The performance difference was stark despite using fairly accurate taggers, which is not a reasonable assumption for low resource languages. Tagging performance in low resource settings is still very weak even when utilising cross-lingual techniques and other forms of weak supervision \citep{kann2020weakly}. Even when more annotated data is available, it isn't clear how useful POS tags are for neural dependency parsers, especially when utilising character embeddings \cite{ballesteros2015improved,de2017raw}. Work investigating the utility of POS tags typically observe a small increase in performance or no impact when used as features for neural dependency parsers. \citet{smith2018investigation} found that universal POS (UPOS) tags offer a marginal improvement for their transition based parser for multi-lingual universal dependency (UD) parsing. \citet{dozat2017stanford} also observed an improvement in parsing performance for graph-based parsers when the predicted UPOS tags came from sufficiently accurate taggers. 

\citet{zhang2020pos} only found POS tags to be useful for English and Chinese when utilising them as an auxiliary
task in a multi-task system. \citet{anderson-gomez-2020-frailty} found that a prohibitively high accuracy was needed to utilise predicted UPOS tags for both graph- and transition-based parsers for UD parsing. They also obtained results that suggested smaller treebanks might be able to directly utilise less accurate UPOS tags. 
We evaluate this further by analysing the impact of tagging accuracy on UD parsing in low resource contexts, 
with regards to the amount of data available to train taggers and parsers. 

\section{Methodology}
We performed three experiments. The first is an evaluation of predicted tags as features for biaffine parsers for real low resource treebanks. It also includes parsers trained with UPOS tagging as an auxiliary task similar to the experiments in \citet{zhang2020pos}. The second experiment evaluates the impact of different tagging accuracies on different dataset sizes using artificial low resource treebanks by sampling from high resource treebanks. The last experiment utilises a data augmentation technique to investigate the efficacy of predicted UPOS tags for very small treebanks ($\sim$20 sentences) when augmented with varying amounts of data.
\paragraph{Low resource data} We take all UD v2.6 treebanks \cite{ud26} with less than 750 sentences in both its training dataset and development dataset. We cluster these treebanks into two groups, very low with less than 50 sentences and low with less than 750. The very low resource treebanks consist of Buryat BDT (bxr), Kazakh KTB (kk), Kurmanji MG (kmr), Livvi KKPP (olo), and Upper Sorbian UFAL (hsb). The low resource set is made up of Belarusian HSE (be), Galician TreeGal (gl), Lithuanian HSE (lt), Marathi UFAL (mr), Old Russian RNC (orv), 
Tamil TTB (ta), and Welsh CCG (cy). We combined the training and development data (when available) to then split them 80$|$20. The statistics for the resulting splits are shown in Table \ref{tab:lr_stats}. We use the original test data for analysis.
\paragraph{Artificial low resource data} We use Indonesian GSD (id), Irish IDT (ga), Japanese GSD (ja), and Wolof WTB (wo) to create artificially low resource treebanks. We take a sample of 100, 232, and 541 sentences from the training and development data. These are then split 80$|$20 for training and development data. We do this three times for each treebank size so we have multiple samples to verify our results. We use the original test data for analysis.
\paragraph{Augmented data} For the experiment using augmented data we use a subset of the smallest treebanks, namely Kazakh, Kurmanji, and Upper Sorbian. We then generate data using the subtree swapping data augmentation technique of \citet{dehouk20}. We generate 10, 25, and 50 trees for each and we then split them 80$|$20. We do this three times for each number of generated trees. We use the original test data for analysis.

\paragraph{Subtree swapping} We gather all the sub-trees with a continuous span which has a \texttt{NOUN}, \texttt{VERB}, \texttt{ADJ} or \texttt{PROPN} as its root node. Other UPOS tags are not used due the likelihood of generating ungrammatical structures. With regards to the permitted relation of the root nodes, we consider all core arguments, all nominal dependents, and most non-core dependents (excluding \texttt{discourse}, \texttt{expl} and \texttt{dislocated}). Then given a tree, we swap one of its sub-trees with one from another tree given that their respective roots have the same UPOS tag, dependency relation and morphological features and given that the sub-trees are lexically different. We repeat the process a second time using a third tree. During this second swap, we do not allow the previously swapped subtree to be altered again so as to avoid redundancy. For a more detailed description of this process see \citet{dehouk20}. We create all possible trees generated from the three original trees given the constraints described above, repeat this for each triplet of trees, and finally take a sample from this set of augmented data.

\begin{table}[htbp!]
    \centering
    \small
    \begin{tabular}{lrrrr}
    \toprule
        & \multicolumn{2}{c}{Train} & \multicolumn{2}{c}{Dev}\\
        & sents & tokens & sents  & tokens\\
    \midrule
    \textbf{bxr} & 15 & 120 & 4 & 33\\
    \textbf{kk}  & 24 & 395 & 7 & 134\\ 
    \textbf{kmr} & 16 & 192 & 4 & 50\\
    \textbf{olo} & 15 & 114 & 4 & 30\\
    \textbf{hsb} & 18 & 310 & 5 & 150\\ \midrule
    \textbf{be}  & 307 & 6,441 & 77 & 1,449\\
    \textbf{gl}  & 480 & 12,317 & 120 & 3,119\\
    \textbf{lt}  & 166 & 3,444 & 42 & 852\\
    \textbf{mr}  & 335 & 2,751 & 84 & 686\\
    \textbf{orv} & 256 & 8,253 & 64 & 1903\\
    \textbf{ta}  & 383 & 6,082 & 96 & 1,254 \\
    \textbf{cy}  & 491 & 10,719 & 123 & 2,616\\\bottomrule
    \end{tabular}
    \caption{Number of trees in training and development splits as used for low resource UD treebanks.}
    \label{tab:lr_stats}
\end{table}
\paragraph{Controlling UPOS accuracy}
For each treebank size and split for the artificial low resource treebanks we trained taggers with varying accuracies (60, 66, 72, 78, 85, 89). We allowed a small window around the accuracy for each bin of $\pm$0.25. Following a similar methodology to \citet{anderson-gomez-2020-frailty} to obtain taggers with varying accuracies, we train the taggers as normal and save models when they reach a desired accuracy. We then train parsers using predicted tags from each of the taggers and use predicted tags at inference. For the data augmentation experiment we used accuracy bins of 41, 44, 48, and 51.
\begin{figure*}[thpb!]
    \centering
    \includegraphics[width=.78\linewidth]{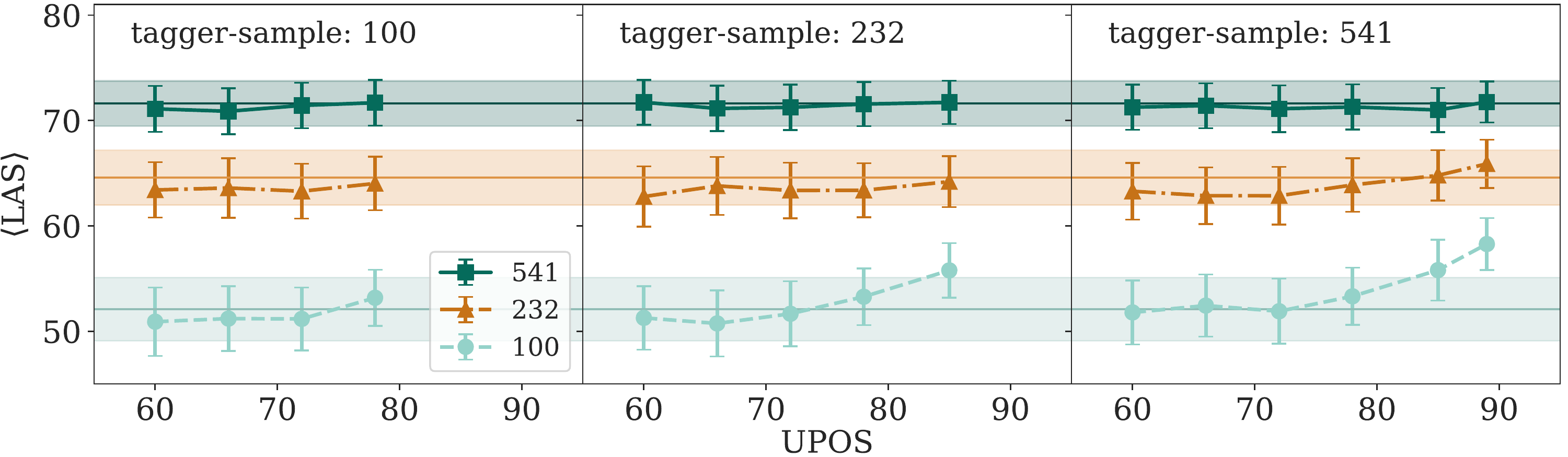}
    \caption{Impact of tagging accuracy for varying amounts of data for both taggers and parsers using artificial low resource data. The standard error of UPOS accuracy is not shown as it is very small ($<0.1\%$ relative error for all bins). Horizontal lines and corresponding shaded area show the mean parsing performance and the standard error for the baseline parsers trained without UPOS tags.}
    \label{fig:artificial-low-results}
\end{figure*}
\paragraph{Network details} Both the taggers and parsers use word embeddings and character embeddings. The parsers use UPOS tag embeddings except for the MTL setup and the baseline models without tags. The embeddings are randomly initialised. The parsers consist of the embedding layer followed by BiLSTM layers and then a biaffine mechanism \cite{dozat20161}. The taggers are similar but with an MLP following the BiLSTMs instead.  We ran a hyperparameter search evaluated on the development data of Irish and Wolof. This resulted in 3 BiLSTM layers with 200 nodes, 100 dimensions for each embedding type with 100 dimensions for input to the character LSTM. The arc MLP of the biaffine structure has 100 dimensions, whereas the relation MLP has 50.

\section{Results and discussion}

Table \ref{tab:low_results} shows the real low resource treebank results. Table \ref{tab:very_low} shows the results for the treebanks with less than 50 sentences. The performance is very low across the board so it is difficult to draw any substantial conclusions, however, using gold tags has a large impact over not using any, almost doubling the labeled attachment score. Also, using predicted tags does result in an increase on average, but Kazakh and Kurmanji lose almost a point. Further those two treebanks and also Buryat have reasonable gains when using the multi-task framework. The average multi-task score is strongly affected by the large drop seen for Upper Sorbian, which also suffers with respect to tagging accuracy when using the multi-task setup. 

\begin{table}[b!]
\centering
\small
\tabcolsep=.18cm    
\begin{subtable}[t]{0.99\columnwidth} 
    \begin{tabular}{lcc|cccc}
    \toprule
         & \multicolumn{2}{c}{UPOS} & \multicolumn{4}{c}{LAS} \\
         & Single & Multi & None & Pred & Gold & Multi \\ 
         \midrule
    \textbf{bxr} & 48.72 & 48.34 & 10.45 & 12.36 & 20.31 & 14.41 \\
    \textbf{kk}  & 53.37 & 52.14 & 22.48 & 21.63 & 36.66 & 23.50 \\
    \textbf{kmr} & 50.56 & 53.73 & 19.16 & 18.31 & 35.54 & 21.58 \\
    \textbf{olo} & 37.84 & 37.37 & \invzero9.74 & 10.89 & 17.54 & \invzero7.59 \\
    \textbf{hsb} & 53.44 & 47.28 & 18.36 & 20.03 & 41.88 & 14.66 \\
    \midrule
    \textbf{avg} & 48.79 &  47.77 & 16.04 & 16.64 & 30.39 & 16.25 \\
    \bottomrule
    \end{tabular}
    \caption{Very low resource: less than 50 sentences.}\label{tab:very_low}
\end{subtable}
\vspace{1em}

\begin{subtable}[t]{0.99\columnwidth}
    \begin{tabular}{lcc|cccc}
    \toprule
         & \multicolumn{2}{c}{UPOS} & \multicolumn{4}{c}{LAS} \\
         & Single & Multi & None & Pred & Gold & Multi \\ 
         \midrule
    \textbf{be}  & 92.82 & 87.29 & 61.82 & 64.91 & 68.87 & 62.28 \\
    \textbf{gl}  & 93.54 & 88.56 & 70.60 & 72.73 & 79.06 & 70.54 \\
    \textbf{lt}  & 79.25 & 71.51 & 37.17 & 35.94 & 48.30 & 38.96 \\
    \textbf{mr}  & 80.58 & 76.46 & 57.04 & 58.74 & 64.32 & 56.31 \\
    \textbf{orv} & 87.77 & 81.60 & 49.53 & 51.34 & 60.24 & 50.33 \\
    \textbf{ta}  & 86.88 & 79.23 & 63.85 & 62.75 & 74.31 & 63.15 \\
    \textbf{cy}  & 91.77 & 86.41 & 72.10 & 72.93 & 80.71 & 73.00 \\
    \midrule
    \textbf{avg} & 85.89 & 77.77 & 55.24 & 56.52 & 64.13 & 55.10 \\
    \bottomrule
    \end{tabular}
    \caption{Low resource: less than 750 sentences.}\label{tab:low}
\end{subtable}
\caption{Performance of different low resource parsers: using predicted UPOS tags as features (Pred), multi-task system where tagging is an auxiliary task to parsing (Multi), using gold UPOS tags as features (Gold), and without using UPOS tags as features (None). The accuracies of the predicted UPOS tags (Single) and that of the multi-task (Multi) are also reported.}\label{tab:low_results}
\end{table}

\begin{figure*}[thpb!]
    \centering
    \includegraphics[width=.74\linewidth]{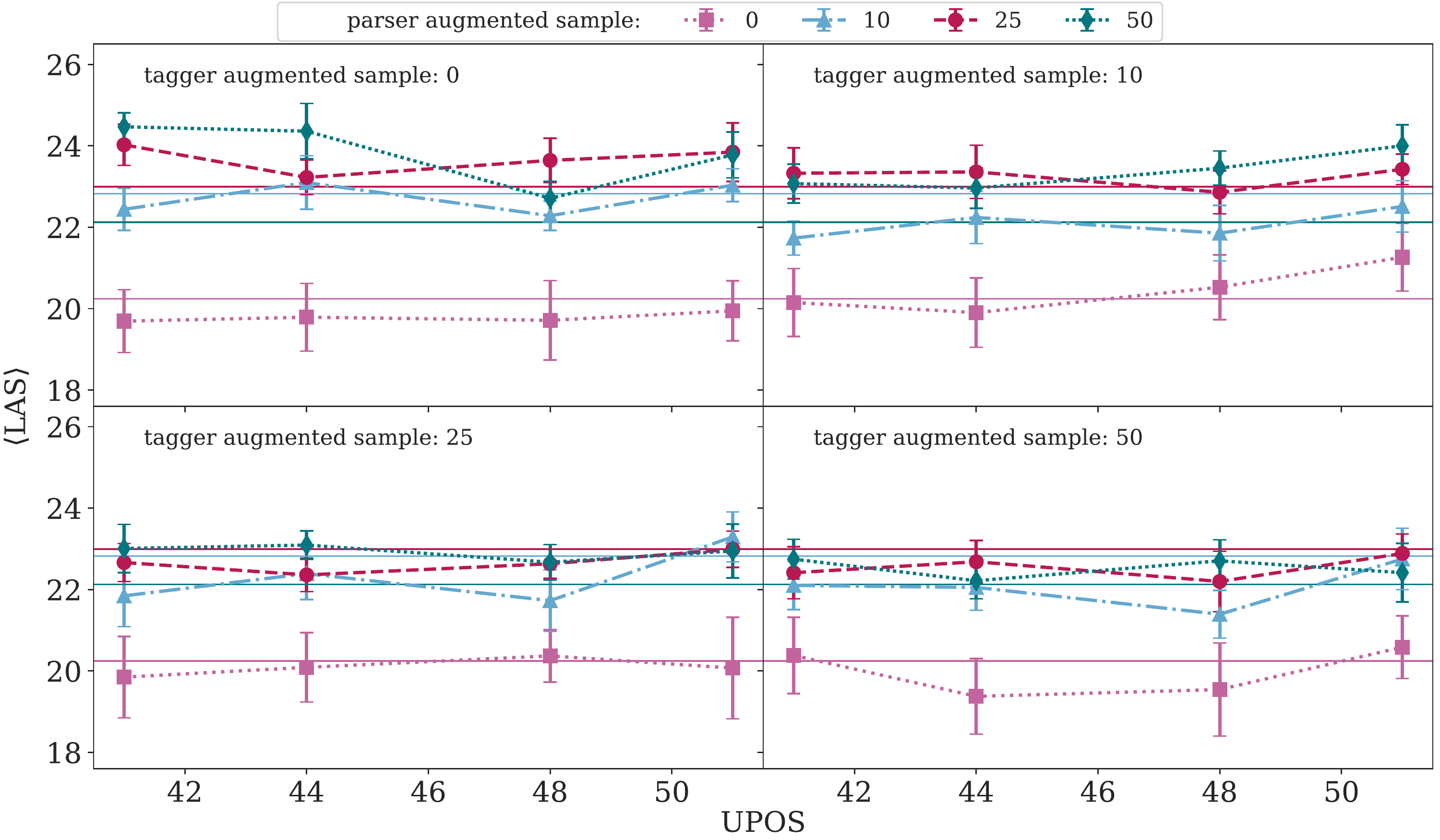}
    \caption{Impact of tagging accuracy for varying amounts of data for both taggers and parsers using augmented data (0, 10, 25, and 50 augmented trees) on top of the original gold data. The standard error of UPOS accuracy is not shown as it is very small ($<0.1\%$ relative error for all bins). Horizontal lines show the mean parsing performance for the baseline parsers trained without UPOS tags (standard error not shown due to too much overlap between augmented data sample sizes).}
    \label{fig:augmented}
\end{figure*}

Table \ref{tab:low} shows the results for the low resource treebanks with less than 750 sentences. On average using predicted UPOS tags achieves a sizeable increase over not using any tags of about 1.2, despite the average tagging accuracy only being 85.89\%. This suggests that in a lower resource setting  the tagging accuracy doesn't have to be quite so high as is needed for high resource settings. Increases in performance are seen for all treebanks except Lithuanian and Tamil. While Lithuanian has the second lowest tagging score, Tamil has a fairly high score, so it seems that the accuracy needed is somewhat language-specific or at the very least data-dependent. The difference for the treebanks in Table \ref{tab:low} is almost 9 points higher for using gold tags. The multi-task performance is about 1.4 points less than using predicted tags on average. However, Lithuanian and Tamil obtain an increase in performance using the multi-task system in comparison to using predicted tags. 

Figure \ref{fig:artificial-low-results} shows the average LAS performance for the parsers trained with the artificial low resource data. When the parsers have sufficient data, using UPOS tags doesn't offer any improvement in performance. For the parsers trained with 232 samples, there is a slight upward trend when using tags predicted from taggers trained with 541 samples. The improvement increases with respect to UPOS tag accuracy and exceeds the performance of the parsers trained with no UPOS tags. The most noticeable improvement is for the parsers trained with only 100 samples. The impact of UPOS accuracy is clearer as the tagger sample size increases as higher accuracies can be obtained. The best performance is with the most accurate taggers (89\%). 

This is a potentially useful finding if annotators have little time, as annotating UPOS tags is much less time-sensitive and can help improve parsing performance if a limited number of tree-annotated sentences are available. However, taking parsers using only 100 fully-annotated training sentences as a baseline, the average performance using 232 parsed sentences without UPOS tags is over 10 points higher, whereas the increase gained training the taggers with 541 tagged sentences is only 5 points. So it is clear that if time permits such that annotators can increase the number of tree annotations, they will likely prove to be more useful. But UPOS tags could be obtained using projection methods and/or active learning techniques \cite{baldridge2009well,das2011unsupervised,garrette2013real,tackstrom2013token}. Also, multilingual projection methods could be used, but they typically generate trees as well as POS tags \cite{agic2016multilingual,johannsen2016joint}.

Figure \ref{fig:augmented} shows the impact of predicted UPOS accuracy when using data generated with subtree swapping augmentation. The first result worth noting is that the augmented data increases performance in this very low resource context. Across the board, the best performing parsers using augmented data outperform the parsers trained only on gold data by 3-6 points which corroborates the findings in previous work. However, it appears that there is a limit to how much augmented data helps as the performance of the parsers which use 25 and 50 augmented instances is similar. 

It also appears that this upper limit is even lower for training taggers with the best performance coming when using predicted tags from taggers utilising only 10 augmented samples or none at all. Using more invariably hurts performance no matter what accuracy the taggers obtained, as can be seen in the subplots showing the performance for parsers trained with predicted tags from taggers using 25 and 50 augmented samples. Also, there is no clear trend showing the impact of UPOS accuracy in this very low resource context. 
\section{Conclusion}
We have presented results which suggest that lower accuracy taggers can still be beneficial when little data is available for training parsers, but this requires a high ratio of UPOS annotated data to tree annotated data.  Experiments using artificial low resource treebanks highlight that this utility diminishes if the number of samples reaches a fairly small amount. We have also shown that very small treebanks can benefit from augmented data and utilise predicted UPOS tags even when they come from taggers with very low accuracy. Our experiments haven't considered pretrained multilingual language models (LMs) which could potentially offset the small benefits of using POS tags. It would be interesting to develop this analysis further by testing whether the implicit information encoded in these LMs are more useful than explicit but potentially erroneous POS tag information. Finally, as one reviewer highlighted, the set of POS tags in the UD framework might just not be sufficiently informative in this setting. While this might be true, the greater contributing factor is surely the low accuracy of the taggers. 
\section*{Acknowledgments}
This work has received funding from the European Research Council (ERC), under the European Union's Horizon 2020 research and innovation programme (FASTPARSE, grant agreement No 714150), from ERDF/MICINN-AEI (ANSWER-ASAP, TIN2017-85160-C2-1-R), from Xunta de Galicia (ED431C 2020/11), and from Centro de Investigación de Galicia ``CITIC'', funded by Xunta de Galicia and the European Union (ERDF - Galicia 2014-2020 Program), by grant ED431G 2019/01.
\bibliography{library}
\bibliographystyle{acl_natbib}
\end{document}